\documentclass[letterpaper, 10 pt, conference]{ieeeconf}

\IEEEoverridecommandlockouts
\usepackage{amsmath,amssymb}
\usepackage{graphicx}
\usepackage{booktabs}
\usepackage{cite}
\usepackage{url}
\usepackage{balance}
\usepackage{tikz}
\usetikzlibrary{arrows.meta,positioning}

\title{\LARGE \bf
Paraphrase-Induced Output-Mode Collapse: When LLMs Break Character Under Semantically Equivalent Inputs
}

\author{Aofan Liu$^{1,\dagger}$ \quad Jingxiang Meng$^{2,\dagger,*}$%
\thanks{$^{1}$Peking University.}%
\thanks{$^{2}$University of Chicago.}%
\thanks{$^{\dagger}$Both authors contributed equally to this work. Author order is alphabetical by last name.}%
\thanks{$^{*}$Corresponding author.}%
}

\begin{document}

\maketitle
\thispagestyle{empty}
\pagestyle{empty}

\begin{abstract}
	When the substantive content of a request is rewritten, do large language models still answer in the format the original task asked for? We find that they often do not, even at temperature zero. On a 150-query evaluation over five compact 2025-era LLMs and four task types, we observe a systematic failure mode we call \emph{prompt-variant output-mode collapse}: when a closed-form prompt asks for a bare label or a single choice token, content-preserving prompt variants can push the model into conversational prose, the requested format dissolves, and exact-match evaluation pipelines silently misjudge the result. To make this measurable, we release \textsc{ParaConsist}, a 900-prompt benchmark of 150 base queries with five lexical, syntactic, and semantic-expansion prompt variants each, and a Semantic Consistency Score that decomposes prompt-variant robustness into answer consistency, sentence-BERT semantic similarity, and length stability. Under a whole-word answer-set match, only $\sim$22\% of closed-form variant responses preserve the ground-truth label inside their output, while $\sim$78\% drift away from the answer space entirely. In our pool, the dominant predictor of collapse is task structure rather than model identity, with model differentiation jointly carried by answer consistency and length stability. Robustness audits should therefore track response-mode preservation as a first-class reliability target alongside answer accuracy.
\end{abstract}

\section{INTRODUCTION}

Large Language Models (LLMs) have demonstrated remarkable capabilities across diverse natural language processing tasks, achieving human-level or near-human performance on numerous benchmarks~\cite{brown2020language,chowdhery2022palm,openai2023gpt4}. However, recent studies have revealed a critical vulnerability: these models exhibit significant brittleness to input variations that preserve the task content while changing its surface realization, producing inconsistent outputs when presented with rewritten prompts~\cite{ribeiro2020beyond,zhu2023promptbench}. This phenomenon, termed \emph{surface-form sensitivity}, poses fundamental challenges for the reliable deployment of LLMs in real-world applications where users naturally express queries in diverse linguistic forms.

The implications of this inconsistency extend beyond mere performance metrics. In conversational AI systems, users frequently reword the substantive content of a task when seeking clarification or exploring different aspects of a topic. When an LLM provides contradictory answers or changes response format under content-preserving rewrites, it erodes user trust and raises serious concerns about the model's reliability~\cite{wang2023selfconsistency,chen2023robustness}. Moreover, this vulnerability can be exploited through adversarial paraphrasing to bypass safety mechanisms~\cite{zou2023universal,wei2023jailbroken,liu2025pico,yang2025multiagent}, highlighting critical security implications for deployed systems.

Despite growing awareness of this issue, the research community lacks a systematic framework for measuring and analyzing LLM self-consistency under content-preserving prompt variants. Existing work has examined specific aspects of prompt sensitivity---including adversarial robustness~\cite{zou2023universal,wei2023jailbroken}, judge reliability in evaluation~\cite{zheng2023judging}, and domain-specific brittleness~\cite{ribeiro2020beyond}---but no comprehensive benchmark exists for quantifying robustness across general-purpose tasks while also tracking whether the requested response mode survives. Current evaluation paradigms focus predominantly on single-prompt accuracy, potentially overestimating model capabilities by failing to capture performance variations under rewritten inputs~\cite{zhu2023promptbench}.

In this work, we introduce \textbf{ParaConsist}, a systematic evaluation framework for measuring LLM self-consistency under content-preserving prompt variants, and use it to characterize a deployment-relevant failure mode: when rewritten task content causes a model to abandon the requested response format and drift into conversational prose.

Our contributions are:

\begin{itemize}
	\item \textbf{We identify and quantify prompt-variant output-mode collapse.} Across all five evaluated models and all four task types, content-preserving prompt variants frequently induce models to ``break character'' and return prose where a label or otherwise constrained response was expected, even at $T{=}0$. This makes response-mode preservation itself a first-class reliability target.

	\item \textbf{Task structure is the strongest predictor of collapse in our measured setup.} Across four task types and five compact API models, the SCS gap between MC-QA ($0.439$) and Classification ($0.269$) exceeds the aggregate spread across the five evaluated models, and task type explains consistency much more strongly than model family. The pooled AC ranking of tasks (MC-QA $0.292$ $>$ Classification $0.240$ $>$ Sentiment $0.136$) preserves the same ordering, so we interpret task structure as the dominant observational effect on collapse risk rather than as a clean causal estimate of any single interface intervention.

	\item \textbf{Component-wise evaluation exposes a hidden failure in exact-match pipelines.} We construct a 900-prompt benchmark (150 base queries $\times$ 6 phrasings) spanning ARC-Challenge, a balanced movie-review set, AG News, and XSum, generating five prompt variants per query via lexical, syntactic, and semantic-expansion transformations, and aggregate outputs into a Semantic Consistency Score (SCS) combining Answer Consistency (whole-word answer-set match), semantic similarity, and Length Stability. The bucket-level analysis (Figure~\ref{fig:ac_zero}) shows that only $\sim$22\% of closed-form variant responses keep the ground-truth label anywhere in the output, exposing a deployment-relevant reliability gap that single-number exact-match evaluations miss. We therefore use SCS as a supporting diagnostic rather than as the primary definition of collapse.

	\item \textbf{Statistically grounded but cautious cross-family comparison.} Controlling for task type, we find significant cross-family but non-significant within-family differences. We interpret this as a pattern consistent with family-level differences in training or alignment, while noting that the narrow compact-model range in our pool limits stronger claims.

	\item \textbf{Release of dataset, prompt variants, per-call traces, and analysis scripts} as anonymized supplementary material, enabling bit-level replay of every reported statistic.
\end{itemize}

\section{RELATED WORK}

\subsection{LLM Robustness and Consistency}

The reliability of LLMs under input variations has emerged as a critical research area. Recent work has examined adversarial robustness through jailbreak attacks~\cite{zou2023universal} and prompt injection vulnerabilities~\cite{perez2022ignore}, demonstrating that carefully crafted perturbations can elicit unintended behaviors. Complementary research has investigated self-consistency in multi-step reasoning~\cite{wang2023selfconsistency}, showing that sampling multiple reasoning paths can improve accuracy but also revealing significant output variance across semantically equivalent prompts.

Surface-form sensitivity in benchmark evaluation has been documented by Sclar et al.~\cite{sclar2023quantifying}, who report that subtle changes to prompt formatting (separators, casing, ordering) can swing few-shot accuracy by up to 76 percentage points on LLaMA-2-13B, though their analysis targets prompt-template perturbations on the evaluator side rather than content-preserving rewrites of the user payload. Ribeiro et al.~\cite{ribeiro2020beyond} demonstrated similar brittleness in NLU models through behavioral testing, but their framework predates modern LLMs.

Judge reliability studies~\cite{zheng2023judging} have examined consistency in LLM-as-judge scenarios, revealing position bias and verbosity preferences. Our work differs by providing a systematic framework for measuring response-mode inconsistencies across diverse task types with multi-dimensional metrics.

\subsection{Prompt Rewriting and Semantic Equivalence}

Paraphrase generation techniques have evolved from rule-based~\cite{madnani2010generating} and statistical methods~\cite{bannard2005paraphrasing} to neural approaches~\cite{prakash2016neural}. Recent work leverages large language models for high-quality paraphrase generation~\cite{jiang2022promptbert}, demonstrating superior semantic preservation. Semantic textual similarity metrics, including embedding-based approaches~\cite{reimers2019sentence} and learned similarity functions~\cite{cer2017semeval}, provide tools for quantifying semantic equivalence. Our setting is adjacent but not identical: the variants preserve the task payload or source content while changing the surrounding prompt realization, including cues that may implicitly specify the expected response mode.

\subsection{LLM Evaluation Methodologies}

Standard benchmarks such as MMLU~\cite{hendrycks2020measuring}, ARC~\cite{clark2018think}, and various generation tasks~\cite{narayan2018don} have become canonical for evaluating LLM capabilities. However, these benchmarks typically employ single-prompt evaluation, potentially overestimating model reliability. Consistency metrics in natural language generation evaluation~\cite{celikyilmaz2020evaluation} have focused primarily on factual consistency and hallucination detection~\cite{maynez2020faithfulness}. Our Semantic Consistency Score (SCS) extends these concepts by incorporating answer stability, semantic similarity, and confidence patterns into a unified metric applicable across task types.

Most directly related are recent works measuring LLM self-consistency under paraphrasing: Elazar et al.~\cite{elazar2021measuring} introduce ParaRel (328 paraphrases for 38 factual relations) for pretrained models; Raj et al.~\cite{raj2022measuring} propose a semantic-consistency metric for generative LLMs on closed-book QA; Mizrahi et al.~\cite{mizrahi2024state} scale multi-prompt evaluation to 5K paraphrases across 20 primarily pre-2025 LLMs on 39 tasks; Errica et al.~\cite{errica2025what} introduce complementary sensitivity and consistency metrics for classification; and Chatterjee et al.~\cite{chatterjee2024posix} propose POSIX, a log-likelihood-based sensitivity index for open-weight models. Sclar et al.~\cite{sclar2023quantifying} quantify phrasing sensitivity in closed-form QA benchmarks. Our ParaConsist contribution is distinguished by (i) evaluating a uniform set of 2025-era commercial API models at $T{=}0$, (ii) combining answer consistency, semantic similarity (sentence-BERT cosine), and length stability into a unified SCS while reporting the components separately, and (iii) making metric-choice and prompt-rendering failures visible across four task types.

\section{METHOD}
\label{sec:method}

\subsection{Problem Formulation}

We formalize the problem of measuring LLM self-consistency under content-preserving prompt variants as follows. Let an original prompt be decomposed as $p=(u,x)$, where $u$ is the task scaffold or response-mode cue (e.g., ``Classify:'' or ``Category:'') and $x$ is the task payload. We construct variants $p_i=(u_i,x_i)$ where $x_i$ preserves the substantive content of $x$, while $u_i$ may differ in surface form or salience. This deliberately tests whether the model preserves the intended response mode when the prompt payload is rewritten; it is not a claim that every $p_i$ is a strict semantic paraphrase of the full prompt $p$. We consider four task types that represent common LLM applications:

\begin{itemize}
	\item \textbf{Multiple-Choice QA}: Questions with discrete answer spaces (A/B/C/D)
	\item \textbf{Sentiment Analysis}: Binary sentiment classification of text
	\item \textbf{Text Classification}: Multi-class topic categorization
	\item \textbf{Summarization}: Concise summary generation from source text
\end{itemize}

\begin{figure*}[t]
	\centering
	\includegraphics[width=\textwidth]{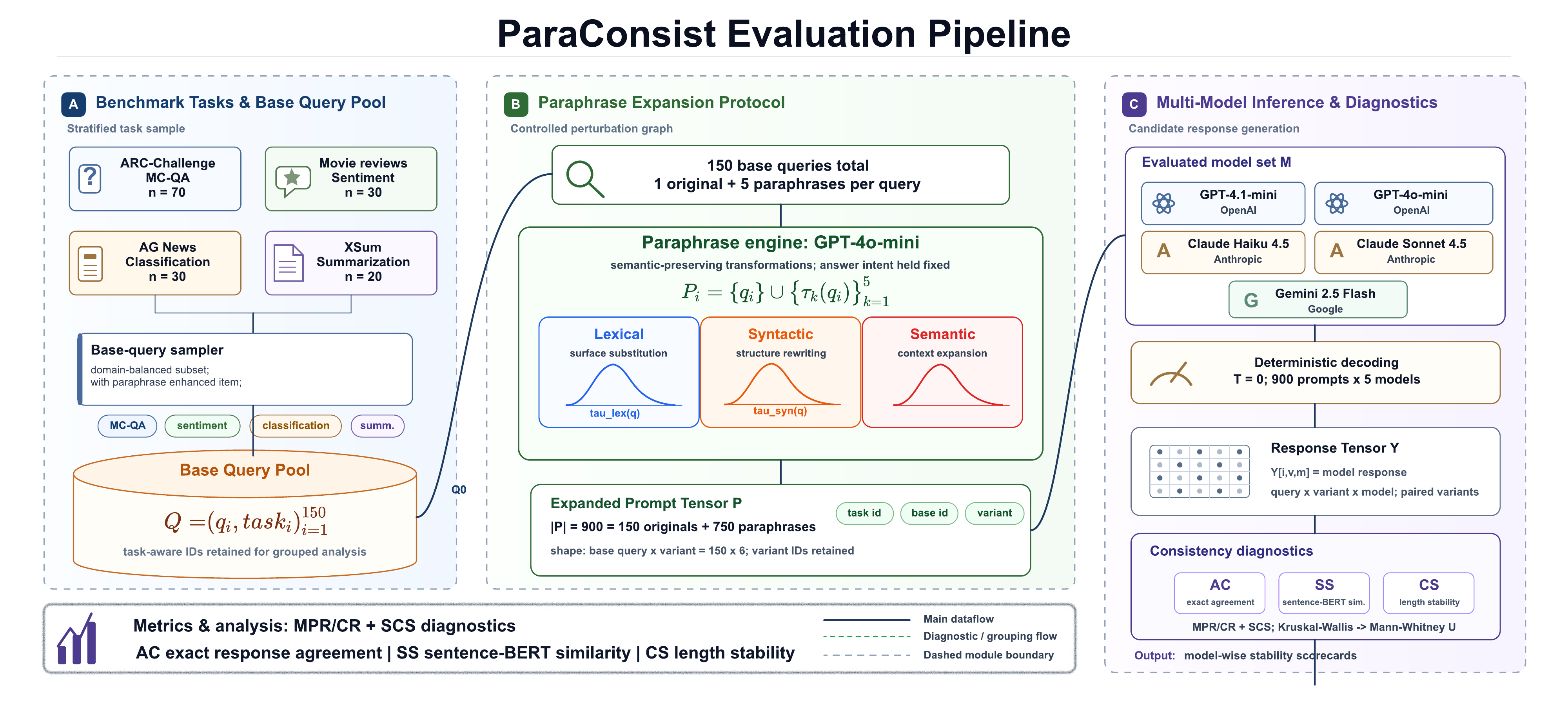}
	\caption{Overview of the \textsc{ParaConsist} pipeline: a 900-prompt benchmark (150 base queries $\times$ 6 prompt renderings) evaluated at $T{=}0$ across five models}
	\label{fig:pipeline}
\end{figure*}

\subsection{Dataset Construction}

Our evaluation framework, \textbf{ParaConsist}, comprises 900 evaluation prompts---150 base queries each rendered as one original phrasing plus five content-preserving variants---spanning four task types. Figure~\ref{fig:pipeline} summarizes the end-to-end evaluation pipeline. The benchmark pool used in this paper is unbalanced because we report the dataset actually used in the current pipeline: Multiple-Choice QA (70 items from ARC-Challenge~\cite{clark2018think}), Sentiment Analysis (30 items from movie review excerpts), Classification (30 items from AG News~\cite{zhang2015character}), and Summarization (20 items from XSum~\cite{narayan2018don}).

For each base item, we generate five content-preserving prompt variants using GPT-4o-mini with carefully designed prompts. The generation strategy targets the task payload rather than exact preservation of the full instruction scaffold, and encompasses three transformation types: (1) lexical substitution---replacing words with synonyms while preserving meaning, (2) syntactic restructuring---altering sentence structure (e.g., active-to-passive voice), and (3) semantic expansion---adding contextual information without changing the core query. This design is intentional: many production failures occur when the user's content is reworded while the expected interface contract remains implicit.

\begin{table}[t]
	\centering
	\small
	\caption{Dataset composition.}
	\label{tab:dataset_stats}
	\begin{tabular}{lrrr}
		\toprule
		\textbf{Task Type} & \textbf{Items} & \textbf{Variants} & \textbf{Total} \\
		\midrule
		Multiple-Choice QA & 70             & 350               & 420            \\
		Sentiment Analysis & 30             & 150               & 180            \\
		Classification     & 30             & 150               & 180            \\
		Summarization      & 20             & 100               & 120            \\
		\midrule
		\textbf{Total}     & 150            & 750               & 900            \\
		\bottomrule
	\end{tabular}
\end{table}

We use all 150 base queries as the primary evaluation set. The resulting run preserves the pipeline's unbalanced task mix---70 MC-QA, 30 sentiment, 30 classification, and 20 summarization items---so we report task-stratified results rather than treating the aggregate as a balanced estimate of population-level consistency for each benchmark family.

\subsection{Consistency Metrics}

We introduce a multi-dimensional \textbf{Semantic Consistency Score (SCS)} that captures three complementary aspects of output stability. For model $M$ on item $q$, let
\[
	\mathbf{c}_{M,q} =
	\left(\text{AC}_{M,q},\,\text{SS}_{M,q},\,\text{CS}_{M,q}\right)
\]
denote the component vector. We aggregate the components with a fixed, answer-prioritized profile:
\begin{equation}
	\begin{aligned}
		\text{SCS}(M)
		 & =
		\frac{1}{|Q|}\sum_{q \in Q}\sum_{r \in \mathcal{R}}
		\pi_r\, g_r(c_{M,q}^{(r)}), \\
		\pi_r
		 & =
		\frac{\exp(\eta_r)}
		{\sum_{s \in \mathcal{R}}\exp(\eta_s)} .
	\end{aligned}
\end{equation}
Here $\mathcal{R}=\{\text{AC},\text{SS},\text{CS}\}$; $g_r(\cdot)$ is the identity on valid component scores and clips numerical edge cases to $[0,1]$; and $\eta$ is a pre-specified component-priority vector satisfying $\eta_{\text{AC}}>\eta_{\text{SS}}>\eta_{\text{CS}}$. In all experiments we use $\pi_{\text{AC}}{=}0.5$, $\pi_{\text{SS}}{=}0.3$, $\pi_{\text{CS}}{=}0.2$, which corresponds to $\eta_{\text{AC}}{-}\eta_{\text{SS}}{=}\log(5/3)$ and $\eta_{\text{SS}}{-}\eta_{\text{CS}}{=}\log(3/2)$ under the softmax parameterization above. The choice puts the largest weight on direct answer agreement (AC) while keeping representational similarity (SS) and surface-form stability (CS) on the same order of magnitude as a tie-breaker between models with similar AC. Thus SCS remains a normalized weighted composite, but the parameterization makes explicit that the composite is a diagnostic profile rather than a learned or universal metric. We report AC, SS, and CS as separate columns in every results table so that readers can reinterpret or re-weight the composite without re-running inference. The components are defined as:

\textbf{Answer Consistency (AC)} measures the agreement rate across prompt variants. For closed-form tasks (QA, classification), we compute exact match ratio:
\begin{equation}
	\text{AC}_{\text{closed}} = \frac{1}{k} \sum_{i=1}^{k} \mathbb{1}[\text{answer}(p_i) = \text{answer}(p)]
\end{equation}
For open-ended generation, we use majority voting on semantic clusters derived from sentence embeddings.

\textbf{Semantic Similarity (SS)} quantifies the semantic coherence of generated outputs using sentence-BERT embeddings~\cite{reimers2019sentence}. We compute pairwise cosine similarities:
\begin{equation}
	\text{SS} = \frac{2}{k(k-1)} \sum_{i=1}^{k-1} \sum_{j=i+1}^{k} \cos(\mathbf{e}_i, \mathbf{e}_j)
\end{equation}
where $\mathbf{e}_i$ is the embedding of the output for variant $p_i$.

\textbf{Length Stability (CS)} measures the consistency of output length across paraphrases, serving as a proxy for response stability:
\begin{equation}
	\text{CS} = 1 - \frac{\sigma(\text{len})}{\mu(\text{len})}
\end{equation}
where $\sigma$ and $\mu$ denote standard deviation and mean of output lengths. A value near 1.0 indicates stable output lengths.

We complement SCS with \textbf{Flip Rate}, the percentage of variants whose answer differs from the original-prompt answer.

\subsection{Evaluation Protocol}

We evaluate five models: GPT-4.1-mini and GPT-4o-mini (OpenAI), Claude Haiku 4.5 and Claude Sonnet 4.5 (Anthropic), and Gemini 2.5 Flash (Google). This selection spans different capability tiers and architectural approaches while remaining within API cost constraints.

To ensure reproducibility, we use deterministic sampling with temperature=0.0. For closed-form tasks, we constrain outputs to valid answer choices. For open generation, we set max\_tokens=512. Each model receives 900 inferences (150 items $\times$ 6 variants: 1 original + 5 prompt variants), totaling 4,500 API calls across all five models.

\section{EXPERIMENTS}

\subsection{Experimental Setup}

We implemented our evaluation framework in Python 3.13, utilizing the LiteLLM proxy for unified model inference and cosine similarity over sentence embeddings for semantic similarity computation. All experiments were conducted on a single machine without GPU requirements, as all inference was performed through API calls.

\subsection{Overall Consistency Scores}

Table~\ref{tab:overall_scs} presents the Semantic Consistency Scores across all five models. All models exhibit moderate-to-low overall SCS values (range: 0.256--0.411), indicating substantial inconsistency when responding to content-preserving prompt variants even under deterministic sampling.

\begin{table}[t]
	\centering
	\small
	\caption{Overall model performance. SCS = Semantic Consistency Score under the fixed answer-prioritized component profile; AC = Answer Consistency; SS = Semantic Similarity; CS = Length Stability.}
	\label{tab:overall_scs}
	\begin{tabular}{lccccc}
		\toprule
		\textbf{Model}    & \textbf{SCS}   & \textbf{AC}    & \textbf{SS} & \textbf{CS} & \textbf{Flip} \\
		\midrule
		GPT-4.1-mini      & 0.262          & 0.153          & 0.257       & 0.541       & 1.00          \\
		GPT-4o-mini       & 0.285          & 0.207          & 0.261       & 0.517       & 1.00          \\
		Claude Haiku 4.5  & 0.344          & 0.273          & 0.209       & 0.723       & 1.00          \\
		Claude Sonnet 4.5 & \textbf{0.411} & \textbf{0.420} & 0.216       & 0.682       & 1.00          \\
		Gemini 2.5 Flash  & 0.256          & 0.060          & 0.255       & 0.749       & 0.99          \\
		\bottomrule
	\end{tabular}
\end{table}

\begin{figure*}[t]
	\centering
	\includegraphics[width=0.95\textwidth]{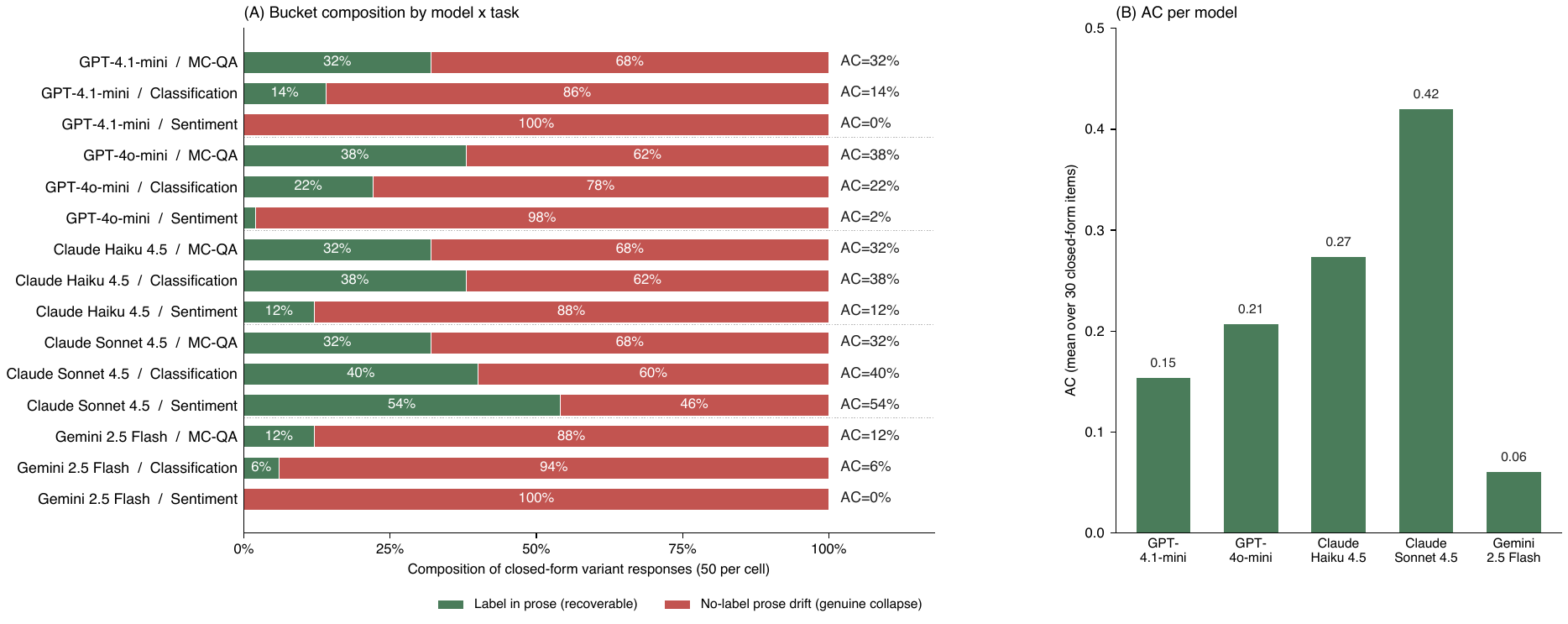}
	\caption{Composition of closed-form variant responses across the five evaluated models. \textbf{(A)} Per (model, task) cell ($50$ variants each), responses are partitioned into \emph{label-in-prose} (green; the ground-truth answer-set token appears as a whole word inside a longer response) and \emph{no-label prose drift} (red; no answer-set token anywhere). Right-margin annotations report the AC, which equals the green share. \textbf{(B)} AC by model on the 30-item closed-form pool, ranging from $0.06$ (Gemini~2.5~Flash) to $0.42$ (Claude~Sonnet~4.5).}
	\label{fig:ac_zero}
\end{figure*}

Across components, confidence/length stability (CS) varies more across model families (Anthropic models 0.682--0.723 and Gemini 0.749 vs.\ OpenAI 0.517--0.541), and semantic similarity (SS) is moderately low across all models (0.209--0.261). Claude~Sonnet~4.5 achieves the highest overall SCS (0.411), driven primarily by its substantially higher AC (0.420), which reflects a high rate of variant responses that contain the ground-truth label embedded in a longer answer (Figure~\ref{fig:ac_zero}A). Claude~Haiku~4.5 follows (0.344), and the OpenAI models cluster in the middle (0.262--0.285). Gemini~2.5~Flash, despite the best CS in the pool, drops to the bottom on SCS (0.256) because almost none of its closed-form variant responses contain the ground-truth label. Model differentiation under prompt rewrites in our pool is therefore carried jointly by AC and CS rather than by either component alone.

\subsection{Task-Type Analysis}
\label{sec:task}

Table~\ref{tab:task_scs} presents the SCS breakdown by task type, averaged across all models. The results reveal substantial variation.

\begin{table}[t]
	\centering
	\small
	\caption{Mean Semantic Consistency Score by task type across all models. The pooled per-task AC values used in this table are: multiple-choice QA $0.292$, classification $0.240$, sentiment $0.136$.}
	\label{tab:task_scs}
	\begin{tabular}{lcccc}
		\toprule
		\textbf{Task Type} & \textbf{Mean SCS} & \textbf{Std} & \textbf{$n$} & \textbf{Rank} \\
		\midrule
		Multiple-choice QA & \textbf{0.439}    & 0.201        & 50           & 1             \\
		Classification     & 0.269             & 0.135        & 50           & 2             \\
		Sentiment          & 0.253             & 0.160        & 50           & 3             \\
		Summarization      & 0.174             & 0.033        & 50           & 4             \\
		\bottomrule
	\end{tabular}
\end{table}

Multiple-choice QA achieves the highest mean SCS ($0.439$), roughly $1.6\times$ that of classification ($0.269$), consistent with the constrained answer space of MC-QA anchoring outputs more tightly than free-form classification labels.

Table~\ref{tab:model_task_matrix} provides the full model $\times$ task interaction matrix, revealing nuanced patterns.

\begin{table}[t]
	\centering
	\small
	\caption{SCS by model and task type. MC-QA leads for four of the five models; Claude~Sonnet~4.5 instead peaks on Sentiment.}
	\label{tab:model_task_matrix}
	\begin{tabular}{lcccc}
		\toprule
		\textbf{Model}    & \textbf{MC-QA} & \textbf{Sent.} & \textbf{Summ.} & \textbf{Class.} \\
		\midrule
		GPT-4.1-mini      & 0.451          & 0.140          & 0.169          & 0.211           \\
		GPT-4o-mini       & \textbf{0.515} & 0.176          & 0.148          & 0.197           \\
		Claude Haiku 4.5  & 0.451          & 0.238          & 0.190          & \textbf{0.360}  \\
		Claude Sonnet 4.5 & 0.445          & \textbf{0.471} & 0.165          & 0.355           \\
		Gemini 2.5 Flash  & 0.334          & 0.242          & \textbf{0.199} & 0.221           \\
		\bottomrule
	\end{tabular}
\end{table}

MC-QA leads for four of the five models, with Claude~Sonnet~4.5 instead peaking on Sentiment ($0.471$, driven by its high sentiment AC of $0.54$). GPT-4o-mini shows the widest within-model spread between its best (MC-QA: $0.515$) and worst (Summarization: $0.148$). By AC, Claude~Sonnet~4.5 leads on sentiment ($0.54$) and classification ($0.40$), while Gemini~2.5~Flash drops to the bottom on both ($0.00$ on sentiment, $0.10$ on classification). The two views together point to highly task-dependent consistency profiles.

\subsection{Statistical Analysis}

\textbf{Model Comparison.} The Kruskal-Wallis test reveals a significant effect of model on SCS ($H = 18.80$, $p < 0.001$), indicating that model choice meaningfully impacts prompt-variant consistency. Pairwise Mann-Whitney U tests (Table~\ref{tab:pairwise}) show that the four OpenAI--Anthropic pairs are significant, while within-family pairs and most Gemini comparisons are not. The AC column in Table~\ref{tab:overall_scs} widens the spread further, with Claude~Sonnet~4.5 ($0.420$) and Gemini~2.5~Flash ($0.060$) bracketing the pool.

\begin{table}[t]
	\centering
	\small
	\caption{Pairwise Mann-Whitney U tests between models on SCS. * denotes $p < 0.05$. OpenAI--Anthropic pairs are significant; within-family pairs and Gemini--Anthropic pairs are not.}
	\label{tab:pairwise}
	\begin{tabular}{lcc}
		\toprule
		\textbf{Model Pair}           & \textbf{$U$} & \textbf{$p$} \\
		\midrule
		GPT-4.1-mini vs GPT-4o-mini   & 749.0        & 0.627        \\
		GPT-4.1-mini vs Claude Haiku  & 498.0        & 0.004*       \\
		GPT-4.1-mini vs Claude Sonnet & 456.0        & 0.001*       \\
		GPT-4.1-mini vs Gemini Flash  & 545.0        & 0.014*       \\
		GPT-4o-mini vs Claude Haiku   & 549.0        & 0.016*       \\
		GPT-4o-mini vs Claude Sonnet  & 501.0        & 0.004*       \\
		GPT-4o-mini vs Gemini Flash   & 610.0        & 0.068        \\
		Claude Haiku vs Claude Sonnet & 741.0        & 0.574        \\
		Claude Haiku vs Gemini Flash  & 855.0        & 0.600        \\
		Claude Sonnet vs Gemini Flash & 991.0        & 0.067        \\
		\bottomrule
	\end{tabular}
\end{table}

\textbf{Task-Type Comparison.} The task-type effect on SCS is highly significant ($H = 73.95$, $p < 10^{-15}$); the pooled per-task AC ordering (MC-QA $0.292$ $>$ classification $0.240$ $>$ sentiment $0.136$) preserves the same ranking. We read it as task-structured observational evidence of collapse risk rather than a clean causal estimate.

\textbf{Effect Size.} Cohen's $d$ between best (Claude~Sonnet~4.5, SCS $0.411$) and worst (Gemini~2.5~Flash, SCS $0.256$) is $\approx\!0.71$ (medium-to-large), with the AC component carrying most of the spread.

\subsection{Qualitative Analysis}
\label{sec:qualitative}

Manual inspection of model outputs reveals several important patterns.

\textbf{Output-mode collapse.} For sentiment and classification, models frequently ``break character'' under content-preserving prompt variants: instead of returning the requested label, they emit conversational prose, recommendations, or follow-up questions. We treat this as \emph{output-mode collapse}: the semantic intent may remain recognizable, but the response mode shifts away from the requested interface. The bucket decomposition in Figure~\ref{fig:ac_zero} quantifies this directly: $\sim$78\% of closed-form variant responses (pooled) contain no answer-set token at all, while the remaining $\sim$22\% retain the correct label inside a longer answer that downstream semantic extraction could still recover.

\subsection{Limitations}

This is a compact evaluation: 900 items per model (70 MC-QA, 30 sentiment, 30 classification, 20 summarization), five compact API models, one prompt-variant generator, leaving open questions about larger pools, broader capability ranges, balanced per-task sampling, and generator-specific bias. Answer Consistency under whole-word matching can over-credit incidental label mentions, so SCS is a supporting diagnostic while the per-(model, task) bucket composition reported in Figure~\ref{fig:ac_zero} remains primary. All runs use temperature$=0.0$; non-zero temperatures would likely amplify the inconsistencies observed.

\section{CONCLUSIONS}

We introduced ParaConsist, a systematic framework for measuring LLM self-consistency under content-preserving prompt variants, and used it to characterize \emph{prompt-variant output-mode collapse} across five compact API models and four task types. All evaluated models remain meaningfully sensitive to prompt rewriting (SCS $0.256$--$0.411$): even at deterministic sampling, $\sim$78\% of closed-form variant responses contain no answer-set token at all (genuine collapse), while $\sim$22\% embed the correct label inside prose (recoverable). Reporting the per-(model, task) bucket composition is necessary to characterize what kind of failure dominates in any given pipeline.

Second, task structure dominates the consistency signal, with the pooled per-task AC ordering MC-QA $>$ classification $>$ sentiment matching the SCS ranking. Per-task reporting and prompt-variant bands are therefore more informative for deployment-oriented evaluation than a single aggregate score.

Third, the cross-family ordering is consistent but the overall-SCS spread is moderate. Claude~Sonnet~4.5 (SCS $0.411$) and Claude~Haiku~4.5 ($0.344$) lead; the OpenAI models cluster in the middle ($0.262$--$0.285$); and Gemini~2.5~Flash ($0.256$) sits at the bottom of the pool. Pairwise tests separate OpenAI from Anthropic, but Gemini--Anthropic differences are not significant on SCS; the cleaner signal sits in AC, where Gemini's $0.060$ trails because almost none of its closed-form variants contain the ground-truth label. Within-family differences remain modest, a pattern consistent with family-level training or alignment differences but plausibly shaped also by the narrow compact-model range in our pool.

Future work should replace exact-match scoring with a semantic answer extractor to quantify how much closed-form failure is recoverable downstream, sweep $T{>}0$ to compound stochasticity with the deterministic inconsistency we already observe, and probe whether the cross-family ranking persists at larger capability tiers.

\balance

\end{document}